\documentclass[10pt,twocolumn,letterpaper]{article}

\usepackage{iccv}
\usepackage{times}
\usepackage{epsfig}
\usepackage{graphicx}
\usepackage{amsmath}
\usepackage{amssymb}

\usepackage[T1]{fontenc}
\usepackage[utf8]{inputenc}
\usepackage{authblk}

\usepackage{iccv}
\usepackage{times}
\usepackage{epsfig}
\usepackage{graphicx}
\usepackage{amsmath}
\usepackage{amssymb}
\usepackage{times}
\usepackage{epsfig}
\usepackage{graphicx}
\usepackage{amsmath}
\usepackage{amssymb}
\usepackage{algorithm} 
\usepackage{algorithmic} 
\usepackage{graphicx,subfigure}
\usepackage{graphics}
\usepackage{multirow}
\usepackage{xcolor}
\usepackage{booktabs}
\usepackage{float} 
\usepackage{stfloats}
\usepackage{flafter}
\usepackage{blindtext}
\usepackage{graphicx}
\usepackage[labelsep=period]{caption}

\makeatletter

\newcommand{\Rmnum}[1]{\expandafter\@slowromancap\romannumeral #1@}
\makeatother

\usepackage[breaklinks=true,bookmarks=false]{hyperref}

\iccvfinalcopy 

\def\iccvPaperID{****} 

\begin{document}

\title{Personalized Age Progression with Aging Dictionary}
\author[$\ddagger$ $\S$ $\dagger$ ]{Xiangbo Shu}
\author[$\ddagger$ $\ast$]{Jinhui  Tang}
\author[$\S$]{Hanjiang  Lai}
\author[$\S$]{Luoqi  Liu}
\author[$\S$]{Shuicheng  Yan}
\affil[$\ddagger$]{School of Computer Science and Engineering, Nanjing University of Science and
	Technology
	}
\affil[$\S$]{Department of Electrical and Computer Engineering, National University of Singapore}
\affil[ ]{\tt\small \{shuxb104,laihanj\}@gmail.com, jinhuitang@njust.edu.cn, \{liuluoqi, eleyans\}@nus.edu.sg}

	\twocolumn[{%
		\renewcommand\twocolumn[1][]{#1}%
		\maketitle
		\begin{center}
					\vspace*{-11mm}
			\centering
			\includegraphics[scale=0.62]{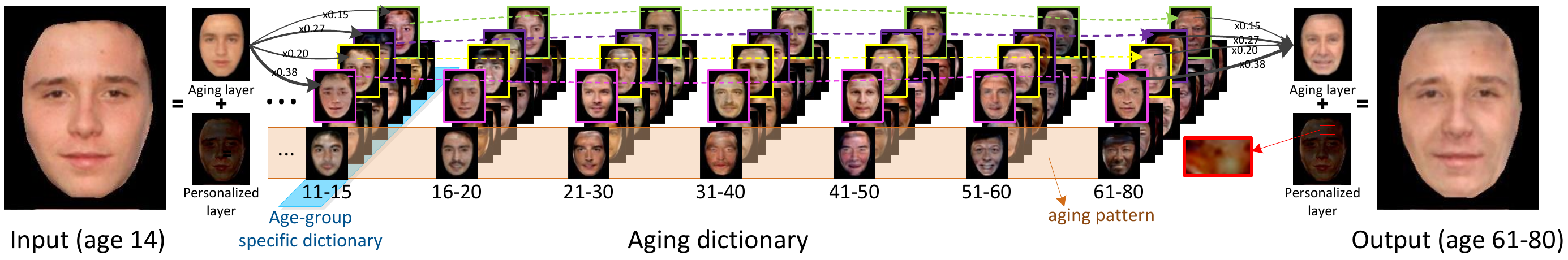}
			\vspace*{-6mm}
			\captionof{figure}{A personalized aging face by the proposed method. The personalized aging face contains the aging layer (e.g., wrinkles) and the personalized layer (e.g., mole). The former can be seen as the corresponding face in a linear combination of the aging patterns, while the latter is invariant in the aging process. For better view, please see $\times3$ original color PDF.}
			\label{fig0}
		\end{center}%
	}]
\maketitle
\renewcommand{\thefootnote}{}
\footnotetext{$\dagger$ This work was performed when X. Shu was visiting National University of Singapore.}
\footnotetext{$\ast$ Corresponding author.}
\renewcommand{\thefootnote}{\arabic{footnote}}
\begin{abstract}
	
		\vspace{-2mm}
	In this paper, we aim to automatically render aging faces in a personalized way. Basically, a set of age-group specific dictionaries are learned, where the dictionary bases corresponding to the same index yet from different dictionaries form a particular aging process pattern cross different age groups, and a linear combination of these patterns expresses a particular personalized aging process. Moreover, two factors are taken into consideration in the dictionary learning process. First, beyond the aging dictionaries, each subject may have extra personalized facial characteristics, e.g. mole, which are invariant in the aging process. Second, it is challenging or even impossible to collect faces of all age groups for a particular subject, yet much easier and more practical to get face pairs from neighboring age groups. Thus a personality-aware coupled reconstruction loss is utilized to learn the dictionaries based on face pairs from neighboring age groups. Extensive experiments well demonstrate the advantages of our proposed solution over other state-of-the-arts  in term of personalized aging progression, as well as the performance gain for cross-age face verification by synthesizing aging faces. 
	
	\vspace{-3mm}

\end{abstract}


\section{Introduction}

Age progression, also called age synthesis~\cite{fu2010age} or
face aging~\cite{suo2010compositional}, is defined as aesthetically rendering a face image with natural aging and rejuvenating effects for an individual face. It has found application in some domains such as cross-age face analysis~\cite{park2010age}, authentication systems, finding lost children, and entertainment. There are two main categories of solutions to the age progression task: prototyping-based age progression~\cite{kemelmacher2014illumination,tiddeman2001prototyping,gandhi2004method} and modeling-based age progression~\cite{suo2010compositional,tazoe2012facial,liang2011multi}. Prototyping-based age progression transfers the differences between two prototypes (e.g., average faces) of the pre-divided source age group and target age group into the input individual face, of which its age belongs to the source age group. Modeling-based age progression models the facial parameters for the shape/texture synthesis with the actual age (range). 

Intuitively, the natural aging process of a specific human usually follows the general rules in the aging process of all humans, but this specific process should also contain some personalized facial characteristics, e.g., mole, birthmark, etc., which are almost invariant with time. Prototyping-based age progression methods cannot well preserve this \textbf{personality} of an individual face, since they are based on the \textit{general} rules in the human aging process for a relatively large population. Modeling-based age progression methods do not specially consider these personalized details. Moreover, they require dense {\bf long-term} (e.g. age span exceeds 20 years) face aging sequences for building the complex models. However, collecting these dense long-term face aging sequences in the real world is very difficult or even unlikely. Fortunately, we have observed that the short-term (e.g. age span smaller than 10 years) face aging sequences are available on the Web, such as photos of celebrities of different ages on Facebook/Twitter. Some available face aging databases~\cite{chen2014cross,fgnet,ricanek2006morph} also contain the dense short-term sequences. Therefore, generating personalized age progression for an individual input by leveraging short-term face aging sequences is more feasible.

In this paper, we propose an age progression method which automatically renders aging faces in a personalized way on a set of age-group specific dictionaries, as shown in Figure~\ref{fig0}.
Primarily, based on the aging-(in)variant patterns in the face aging process, an individual face can be decomposed into an aging layer and a personalized layer. The former shows the general aging characteristics (e.g., wrinkles), while the latter shows some personalized facial characteristics (e.g., mole). For different human age groups (e.g., 11-15, 16-20, ...), we design corresponding aging dictionaries to characterize the human aging patterns, where the dictionary bases with the same index yet from different aging dictionaries form a particular aging process pattern (e.g., they are linked by a dotted line in Figure \ref{fig0}). Therefore, the aging layer of the aging face can be represented by a linear combination of these patterns with a sparse coefficient (e.g., $[0,0.38,0,0,0.20,\cdots]$), where the redundancy between the aging layer and the input face can be defined as the personalized layer, which is invariant in the aging process. The motivation for the sparsity is to use fewer dictionary bases for reconstruction such that the reconstructed aging layer of face can be shaper and less blurred. Finally, we render the aging face in the future age range for the individual input by synthesizing the represented aging layer in this age range and the personalized layer.

\begin{figure*}[t]
	\centering
	\includegraphics[scale=0.44595]{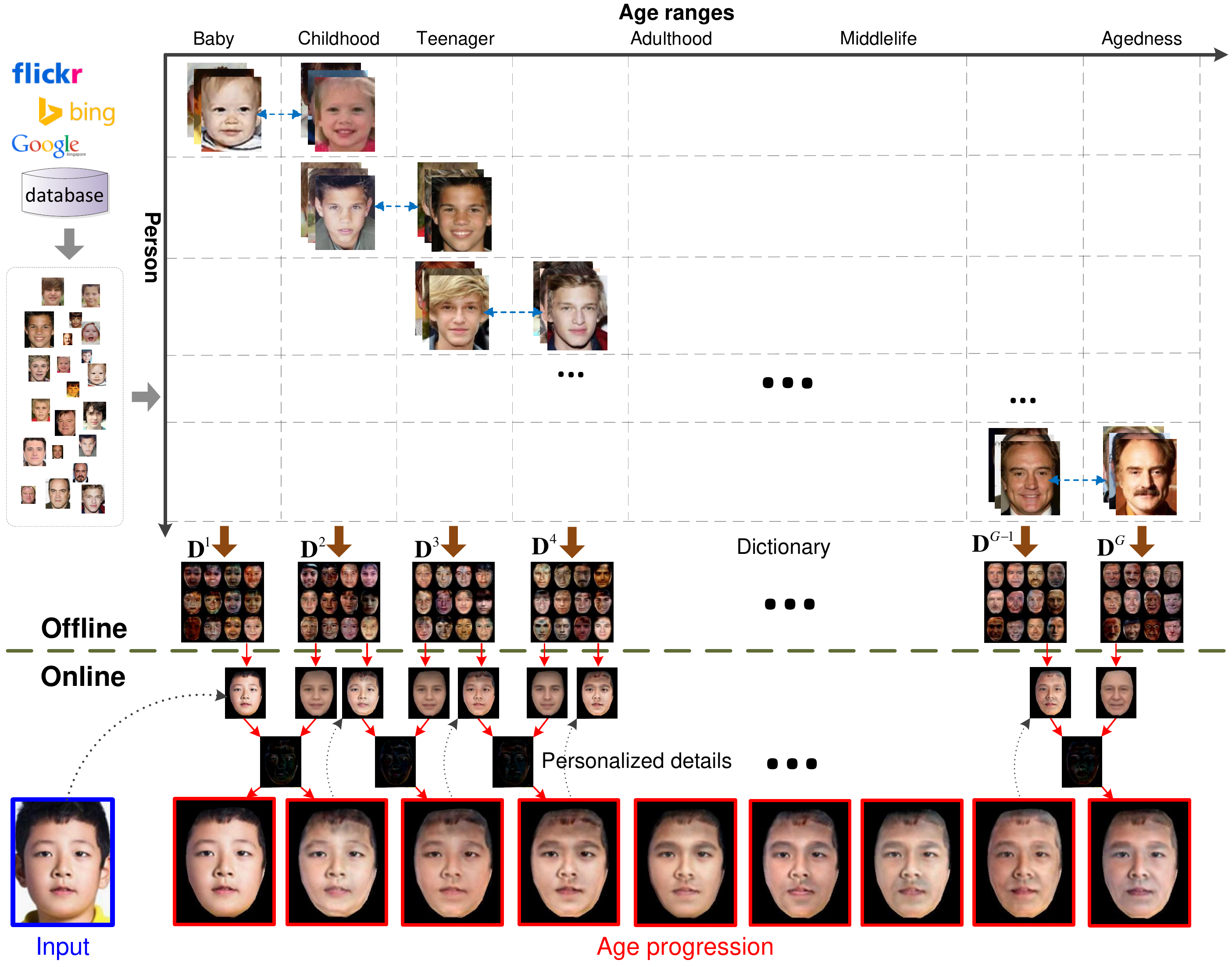}
	\vspace*{-2mm}
	\caption{Framework of the proposed age progression. ${\bf D}^{g}$ denotes a aging dictionary of the age group $g$. In the offline phase, we collect short-term aging face pairs and then train the aging dictionary. In the online phase, for an input face, we firstly render its aging face in the nearest neighboring age group. Taking this aging face as the input of the aging face in the next age group, we repeat this process until all aging faces are rendered. For better view, please see $\times3$ original color PDF.} 
	\label{fig1}
	\vspace*{-3mm}
\end{figure*}

To learn a set of aging dictionaries, we use the more practical short-term face aging pairs as the training set instead of the possibly unavailable long-term face aging sequences. Based on the aging relationships between a face aging pair of the same person covering two neighboring age groups, we assume that the sparse representation of a younger-age face w.r.t. the younger-aging dictionary can represent its older-age face w.r.t. the older-aging dictionary, excluding the personalized layer. The distribution of face aging pairs is shown in the upper part of Figure~\ref{fig1}. We can see that: (1)~each age group has its own aging dictionary, and every two neighboring age groups are linked by the collected dense short-term face aging pairs; (2)~one particular may appear in two different neighboring-group face pairs, which makes all the age groups linked together; (3)~the personalized details (the personalized layer) contain the personalized facial characteristics. These three properties are able to guarantee that all aging dictionaries can be simultaneously trained well by a personality-aware coupled reconstruction loss on the short-term face aging pairs. 

Our main contributions in this paper are two-fold: (1)~we propose a personalized age progression method to render aging faces, which can preserve the personalized facial characteristics; (2)~since it is challenging or even impossible to collect intra-person face sequences of all age groups, the proposed method only requires the available short-term face aging pairs to learn all aging dictionary bases of human aging, which is more feasible. Extensive experiments well validate the advantage of our proposed solution over other state-of-the-arts w.r.t. personalized aging progression, as well as the performance gain for cross-age face verification by synthesizing aging faces. 

\section{Related Work}

Age progression has been comprehensively reviewed in literature~\cite{fu2010age,ramanathan2009age,ramanathan2009computational}.
As one of the early studies, Burt \textit{et al.}~\cite{burt1995perception} focused on creating average faces for different ages and transferring the facial differences between the average faces into the input face. This method gave an insight into the age progression task. Thereafter, some prototyping-based aging methods~\cite{tiddeman2001prototyping,kemelmacher2014illumination} were proposed with different degrees of improvements. In particular, the paper~\cite{kemelmacher2014illumination} leveraged the difference between the warped average faces (instead of the original average faces) based on the flow from average face to input face. A drawback of these methods is that the aging speed for each human is synchronous and no personalized characteristic is saved, which leads to similar aging results of many faces to each other. Although some researchers target at individual-specific face aging~\cite{solomon2006person,hubball2008image,hunter2009synthesis}, lack of personality for aging faces is still a challenging problem.

Modeling-based age progression which considers shape and texture synthesis simultaneously is another popular idea~\cite{albert2007review}. There have been quite a quantity of modeling-based age progression methods proposed, including active appearance model~\cite{lanitis2002toward}, craniofacial growth model~\cite{ramanathan2006face}, and-or graph model~\cite{suo2010compositional}, statistical model~\cite{paysan2010statistical} and implicit function~\cite{berg2006facial,schroeder2007facial}, etc. Generally, to model large appearance changes over a long-term face aging sequence, modeling-based age progression requires sufficient training data. Suo {\em et al.}~\cite{suo2012concatenational} attempted to learn long-term aging patterns from available short-term aging databases by a proposed concatenational graph evolution aging model.

\vspace{0mm}
\section{Personalized Age Progression with Aging Dictionary}

\subsection{Overview of Our Framework}   
The framework of the proposed personalized age progression is plotted in Figure~\ref{fig1}. The offline phase is described as follows. First, we collect the dense short-term aging pairs of the same persons from the Web and also from available databases. Second, for each age group, we design a corresponding aging dictionary to represent its aging characteristics. Third, all aging dictionaries are simultaneously trained by a personality-aware coupled dictionary learning model on the collected database. In the online phase, for an input face, we first construct the aging face in the nearest neighboring age group by the corresponding aging dictionary with an implicitly common coefficient, as well as a personalized layer. After that, taking this new aging face as the input of aging synthesis in the next age group, we repeat this process until all aging faces have been rendered. More details will be described in Section~\ref{APS}. 

\vspace{-1mm}
\subsection{Formulation}
\vspace{-1mm}
We divide the human aging process into $G$ age groups (each group spans less than 10 years) in this paper. Let $\{\textbf{x}_i^1,\cdots,\textbf{x}_i^g,\cdots,\textbf{x}_i^G\}$ denote a selected face aging sequence of the person $i$, where the face photo $\textbf{x}_i^g\in \mathbb{R}^f$ falls into the age group $g$ ($f$ is the number of pixels in the face photo). Assume we have $L$ face aging sequences $\{\textbf{x}_i^1,\textbf{x}_i^2,\cdots,\textbf{x}_i^G\}_{i=1}^L$ in total. For the age group $g$ ($g=1,2,\cdots,G$), we define its aging dictionary ${\bf B}^g$ to capture the aging characteristics, which will be learned in the following. 

{\bf Personality-aware formulation.} Our aging dictionary learning model considers the personalized details of an individual when representing the face aging sequences on their own aging dictionaries. Since the personalized characteristics are aging-irrelevant and -invariant, such as mole, birthmark, permanent scar, etc., we plan to add a personalized layer  ${\bf p}_i\in \mathbb{R}^{f}$ for a face aging sequence $\{\textbf{x}_i^1,\textbf{x}_i^2,\cdots,\textbf{x}_i^G\}$ to indicate the personalized details in the human aging process. Moreover, considering the computational efficiency, we employ PCA projection to reduce the dimension of the dictionary. Let ${\bf H}^g\in \mathbb{R}^ {f \times m}$ denote the PCA projected matrix of all data in the age group $g$, and the original aging dictionary ${\bf B}^g$ is redefined as ${\bf D}^g\in \mathbb{R}^ {m \times k}$, where $k$ is the number of dictionary bases. All aging dictionaries compose an overall aging dictionary ${\bf D}=[{\bf D}^1,{\bf D}^2,\cdots,{\bf D}^G]\in \mathbb{R}^{m\times K}$, where $K=k\times  G$. So far, the aging face $\textbf{x}_i^{g+j}$ of $\textbf{x}_i^g$ 
equals the linearly weighted combination of the aging dictionary bases in the age group $g+j$ and the personalized layer ${\bf p}_i$, i.e., ${\bf x}_i^{g+j}\approx{\bf H}^{g+j}{\bf D}^{g+j}{\bf a}_i+{\bf p}_i$ for $j=1,\cdots, G-g$, where ${\bf a}_i$ and ${\bf p}_i$ are the common sparse coefficient and the personalized layer, respectively. For $L$ face aging sequences $\{\textbf{x}_i^1,\cdots,\textbf{x}_i^G\}_{i=1}^{L}$ covering all age groups, a personality-aware dictionary learning model is formulated as follows,
\begin{equation} \label{eq1}
\begin{aligned}
\vspace{-1mm}
&\mathop {\min }\limits_{\scriptstyle ~{\{\bf D}^{g}\}_{g=1}^G,\hfill \atop
	\scriptstyle {\{{\bf{a}}_i,{\bf{p}}_i\}_{i=1}^L} \hfill} \!\!\sum\limits_{g = 1}^{G} \!\sum\limits_{i = 1}^{L} \!\!\left\{ \!{\left\| {{{\bf{x}}_i^g} \!-\!\!{{\bf{H}}^g} {{\bf{D}}^g}{{\bf{a}}_i} \!\!-\!\! {{\bf{p}}_i}} \right\|_2^2 \!\!+ \!\!  \gamma\! \left\| {{{\bf{p}}_i}} \right\|_2^2\!+\! \lambda {{\left\| {{{\bf{a}}_i}} \right\|}_1}  \!}\!\right\}\\
& ~~~~~\text{s.t.}{\kern 2pt}  {\left\| {{{\bf{D}}^g(:,d)}} \right\|_2} \le 1,\forall  d \!\in\! \{ 1, \cdots ,k\},\forall  g \!\in\! \{ 1, \cdots ,G\},
\end{aligned}
\end{equation}
where ${\bf D}^{g}(:,d)$ denotes the $d$-th column (base) of ${\bf D}^g$, and parameters $\lambda$ and $\gamma$ control the sparsity penalty and regularization term, respectively. ${\bf D}^{g}(:,d)$ is used to represent the specific aging characteristics in the age group $g$.

{\bf Short-term coupled learning.} We have observed that one person always has the dense short-term face aging photos, but no long-term face aging photos. Collecting these long-term dense face aging sequences in the real world is very difficult or even unlikely. Therefore, we have to use the shot-term face aging pairs instead of the long-term face sequences. Let $\textbf{x}_i^g \in \mathbb{R}^{f}$ denote the $\textit{i}$-th face in the age group $g$, and $\textbf{y}_i^{g}\in \mathbb{R}^{f}$ denote the $\textit{i}$-th face of the same person in the age group $g+\!1$, where $g=1,2,\cdots,G-1$. Let every two neighboring age groups share $n$ face pairs, and then there are $N=n\times (G-1)$ face aging pairs in total. For the face aging pairs $\{\textbf{x}_i^g,\textbf{y}_i^g\}_{i=1}^n$ covering the age group $g$ and $g\!+\!1$ ($g=1,2,\cdots,G-1$), we reformulate a personality-aware coupled dictionary learning model to simultaneously learn all aging dictionaries, i.e., 
\begin{equation} \label{eq2}
\begin{aligned}
\mathop {\min }\limits_{\scriptstyle ~~~~~~\{{\bf D}^{g}\}_{g=1}^G,\hfill \atop
	\scriptstyle {\{\{{\bf{a}}_i^{g},{\bf{p}}_i^{g}\}_{i=1}^n}\}_{g=1}^{G-1} \hfill} \!\!&\sum\limits_{g = 1}^{G - 1} \sum\limits_{i = 1}^{n} \!\left\{ \!{\left\| {{{\bf{x}}_i^g} \!-\!{{\bf{H}}^g} {{\bf{D}}^g}{{\bf{a}}_i^g} \!-\! {{\bf{p}}_i^g}} \right\|_2^2 \!\!+ \!\!  \gamma \left\| {{{\bf{p}}_i^g}} \right\|_2^2  }\right.\\
&\left. +{\left\| {{{\bf{y}}_i^g} \!-\! {{\bf{H}}^{g + 1}}{{\bf{D}}^{g + 1}}{{\bf{a}}_i^g} \!-\! {{\bf{p}}_i^g}} \right\|_2^2  \!+\! \lambda {{\left\| {{{\bf{a}}_i^g}} \right\|}_1}}\!\right\}  \\
\text{s.t.}{\kern 2pt}    || {\bf{D}}^g (:,d) &||_2 \le 1,\forall  d \!\in\! \{ 1, \cdots ,k\},\forall  g \!\in\! \{ 1, \cdots ,G\}.
\end{aligned}
\end{equation}

In Eqn.~\eqref{eq2}, every two neighboring aging dictionaries ${\bf D}^g$ and ${\bf D}^{g+1}$ corresponding to two age groups are implicitly coupled via the common reconstruction coefficient ${{\bf{a}}_i^g}$, and the personalized layer ${{\bf{p}}_i^g}$ is to capture the personalized details of the person $i$, who has the face pair $\{\textbf{x}_i^g,\textbf{y}_i^g\}$. It is noted that face aging pairs are overlapped, which guarantees that we can train all aging dictionaries within a unified formulation.
Let ${\bf D}\!=\![{\bf D}^1,\cdots,{\bf D}^G]\in \mathbb{R}^{m \times K}$, ${\bf P}^g=[{\bf p}_1^g,\cdots,{\bf p}_n^g]\in \mathbb{R}^{f\times n}$, ${\bf P}\!=\![{\bf P}^1,\cdots,{\bf P}^{G-1}]\! \in\! \mathbb{R}^{f\times N}$, ${\bf A}^g\!=\![{\bf a}_1^g,\cdots,{\bf a}_n^g]\in \mathbb{R}^{k\times n}$ and ${\bf A}\!=\![{\bf A}^1,\cdots,{\bf A}^{G-1}]\in \mathbb{R}^{K\times N}$, and Eqn.~\eqref{eq2} can be rewritten in the matrix form after some algebratic steps 
\vspace{-2mm}
\begin{equation} \label{eq3}
\begin{aligned}
\mathop {\min }\limits_{{\bf{D}},{\bf{A}},{\bf{P}}}&\sum\limits_{g = 1}^{G - 1} \left\{ {\left\| {{{\bf{X}}^g} -{{\bf{H}}^g} {{\bf{D}}^g}{{\bf{A}}^g} - {{\bf{P}}^g}} \right\|_F^2 +   \gamma \left\| {{{\bf{P}}^g}} \right\|_F^2  }\right.\\
&\left. +{\left\| {{{\bf{Y}}^g} - {{\bf{H}}^{g \!+\! 1}}{{\bf{D}}^{g \!+\! 1}}{{\bf{A}}^g} - {{\bf{P}}^g}} \right\|_F^2  + \lambda {{\left\| {{{\bf{A}}^g}} \right\|}_1}}\right\}  \\
\text{s.t.}~{\kern 4pt} || &{{\bf D}}^g(:,d) ||_2 \le 1,\forall  d \!\in\! \{ 1, \!\cdots\!,\!k\},\forall  g \!\in\! \{ 1, \cdots \!,G\},
\end{aligned}
\end{equation}
where $||{\bf A}^g||_1=\sum\nolimits_{i = 1}^n {||{{\bf a}}_i^g|{|_1}} $.

\subsection{Optimization Procedure}
\label{OP}
The objective function in Eqn.~\eqref{eq3} is convex w.r.t. $\bf{A}$, $\bf{D}$ and $\bf{P}$ separately, which can be iteratively solved through three alternating sub-procedures of optimization. Specifically, we fix the other unknown variables when updating one unknown variable. This process iterates until convergence. 
The iteration steps shall end when the relative cost of the objective function stays unchanged. 
The proposed aging dictionary learning is summarized in Algorithm~\ref{alg1}.

{\bf Updating $\textbf{A}$}.
When updating $\textbf{A}$, we fix \textbf{D} and \textbf{P} and select the terms related to
$\textbf{A}$ in Eqn.~\eqref{eq3}:
\vspace{-2mm}
\begin{equation} \label{eq4}
\begin{aligned}
\mathop {\min }\limits_{{\bf{A}}}~& \sum\limits_{g = 1}^{G - 1} \left\{ {\left\| {{{\bf{X}}^g} - {{\bf{W}}^g}{{\bf{A}}^g} - {{\bf{P}}^g}} \right\|_F^2 }\right. \\
&\left.+\left\| {{{\bf{Y}}^g} - {{\bf{W}}^{g+1}}{{\bf{A}}^g} - {{\bf{P}}^g}} \right\|_F^2 + \lambda {{\left\| {{{\bf{A}}^g}} \right\|}_1} \right\},
\end{aligned}
\end{equation}
where ${{\bf{W}}^g}={{\bf{H}}^g}{{\bf{D}}^g}$ and ${{\bf{W}}^{g+1}}={{\bf{H}}^{g+1}}{{\bf{D}}^{g+1}}$. Let $\textbf{U}^g=\textbf{X}^g-\textbf{P}^g$, $\textbf{V}^g=\textbf{Y}^g-\textbf{P}^g$, and then we also have
\begin{equation} \label{eq5}
\mathop {\min }\limits_{\textbf{A}} \sum\limits_{g = 1}^{G - 1} \!{\left\{ {\left\| {\!\left[\! {\begin{array}{*{20}{c}}
				{{{\bf{U}}^g}}\\
				{{{\bf{V}}^g}}
				\end{array}} \!\right] \!-\! \left[\! {\begin{array}{*{20}{c}}
				{{{\bf{W}}^g}}\\
				{{{\bf{W}}^{g\!+\!1}}}
				\end{array}} \!\right]\!\!{{\bf{A}}^g}} \right\|_F^2 \!\!+\!\! \lambda {{\left\| {{{\bf{A}}^g}} \right\|}_1}} \!\right\}} .
\end{equation}
Eqn.~\eqref{eq5} drops into a classical \textit{Lasso} problem, which is effectively solved by the SPAMS toolbox\footnote{http://spams-devel.gforge.inria.fr/} in this paper.

{\bf Updating $\textbf{P}$.}
\label{UP}
By fixing $\bf{D}$, $\bf{A}$, and omitting the unrelated terms of Eqn.~\eqref{eq3}, we have the objective function w.r.t. $\bf{P}$:
\begin{equation} \label{eq6}
\begin{aligned}
\mathop {\min }\limits_{{\bf P}}\sum\limits_{g = 1}^{G - 1} \left\{ \left\| {{{\bf{Z}}^g} \!-\! {{\bf{P}}^g}} \right\|_F^2 
\!+\!\left\| {{{{\bf R}}^g} \!-\! {{\bf{P}}^g}} \right\|_F^2 \!+\! \gamma\! \left\| {{{\bf{P}}^g}} \right\|_F^2 \!\right\}\!,
\end{aligned}
\end{equation}
where ${\bf Z}^g={{{\bf{X}}^g} - {{\bf{W}}^g}{{\bf{A}}^g}}$, and ${\bf R}^g={{{\bf{Y}}^g} - {{\bf{W}}^{g+1}}{{\bf{A}}^g}}$. For $g=1,2,\cdots, G-1$, solving Eqn.~\eqref{eq6}, and we obtain the updating way of ${\bf P}=[{\bf P}^1,\cdots,{\bf P}^g,\cdots{\bf P}^{G-1}]$ as follows,
\begin{equation} \label{eq7}
{{{\bf P}}^g} = ({{\bf{Z}}^g} + {{\bf{R}}^g})/({2 + \gamma }).
\end{equation}

\begin{algorithm}[t]
	\scriptsize{
		\renewcommand{\algorithmicrequire}{\textbf{Input:}}
		\renewcommand\algorithmicensure {\textbf{Output:} }
		\caption{Aging Dictionary Learning (Offline)}
		\small
		\label{alg1} 
		\begin{algorithmic}[1]
			\REQUIRE  
			{Face aging pairs $\{{\bf X}^g,{\bf Y}^g\}_{g=1}^{G-1}$, $k$, $\lambda$ and $\gamma$.}
			\ENSURE{${\bf D}=[{\bf D}^1,{\bf D}^2,\cdots,{\bf D}^c,\cdots,{\bf D}^G]$.}
			\renewcommand{\algorithmicrequire}{\textbf{Initialization:}} 
			\REQUIRE { ${\bf D}\!=\![{\bf D}^1,\cdots,{\bf D}^c,\cdots,{\bf D}^G]$, ${\bf A}\!=\![{\bf A}^1,\cdots,{\bf A}^g,$\\$\cdots,{\bf A}^{G-1}]$, ${\bf P}\!=\![{\bf P}^1,\cdots,{\bf P}^g,\cdots,{\bf P}^{G-1}]$, and $iter\leftarrow1$.}
			\STATE Normalize \!$\{{\bf X}^g\!,\!{\bf Y}^g\}_{g=1}^{G-1}$ \!and calculate the projected matrices ${\bf H}^1,\cdots,{\bf H}^c,\cdots,{\bf H}^G$.
			\REPEAT
			\FOR{$g = 1,2,\cdots, G-1$}
			\STATE Update ${\bf A}^g$ with Eqn.~\eqref{eq5}.
			\ENDFOR
			\FOR{$c = 1,2,\cdots, G$}
			\STATE Update ${\bf D}^c$ with Eqn.~\eqref{eq9}.
			\STATE Project the columns of $\textbf{D}^c$ onto the unit ball.
			\ENDFOR
			\FOR{$g = 1,2,\cdots, G-1$} 
			\STATE Update ${\bf P}^g$ with Eqn.~\eqref{eq7}.
			\ENDFOR
			\STATE $iter \leftarrow iter+1$.
			\UNTIL{Convergence}
		\end{algorithmic}
	}
\end{algorithm}

{\bf Updating $\textbf{D}$.}
We update $\bf{D}$ by fixing $\bf{A}$ and $\bf{P}$. Specifically,
we update ${{\bf{D}}^c}~(c=1,2,\cdots, G)$ while fixing all the remaining dictionaries excluding ${{\bf{D}}^c}$. We omit the terms which are independent of ${{\bf{D}}}$ in Eqn.~\eqref{eq3}:
\vspace{-2mm}
\begin{equation} \label{eq8}
\begin{aligned}
\!\mathop {\min }\limits_{\{{\bf D}^c\}_{c=1}^G} \!\sum\limits_{c = 1}^{G - 1} \!\left\{ {|| {{{{\bf U}}^c} \!-\! {{\bf{H}}^{c}}{{\bf{D}}^{c}}{{\bf{A}}^c}} ||_F^2 \!+\! \left\| {{{\bf{V}}^c} \!-\! {{\bf{H}}^{c+1}}{{\bf{D}}^{c+1}}{{\bf{A}}^c}} \right\|_F^2} \right\},
\end{aligned}
\end{equation}
where ${\bf U}^c={\bf X}^c-{\bf P}^c$, and ${\bf V}^c={\bf Y}^c-{\bf P}^c$. Solving Eqn.~\eqref{eq8}, and we can obtain a closed-form solution of ${\bf D}^c$.
\begin{equation} \label{eq9}
\begin{aligned}
&{{\bf{D}}^c}\left( \epsilon_1{{{\bf{A}}^{c - 1}}{{({{\bf{A}}^{c - 1}})}^T} +\epsilon_2 {{\bf{A}}^c}{{({{\bf{A}}^c})}^T}} \right) \\
&- ({{\bf{H}}^c})^T\left( {\epsilon_1{{\bf{V}}^{c - 1}}{{({{\bf{A}}^{c - 1}})}^T} +\epsilon_2 {{\bf{U}}^c}{{({{\bf{A}}^c})}^T}} \right) = 0,
\end{aligned}
\end{equation}
where two indicators $\epsilon_1$ and $\epsilon_2$ are defined as follows,
\vspace{-2mm}
\begin{equation} 
\label{eq10}
\left\{ {\begin{array}{*{20}{l}}
	{{\epsilon _1} = 0, {\epsilon _2} = 1,{\kern 2pt}  {\kern 1pt} {\kern 1pt}  {\rm{for}}{\kern 1pt} {\kern 1pt} {\kern 1pt} c = 1;}\\
	{{\epsilon _1} = 1,{\epsilon _2} = 1, {\kern 1pt} {\kern 1pt} {\kern 1pt} {\kern 1pt}  {\rm{for}}{\kern 1pt} {\kern 1pt} {\kern 1pt} c = 2,3, \cdots ,G - 1;}\\
	{{\epsilon _1} = 1,{\epsilon _2} = 0,{\kern 2pt} {\kern 2pt}  {\rm{for}}{\kern 1pt} {\kern 1pt} {\kern 1pt} c = G.}
	\end{array}} \right.
\end{equation}

\vspace{-3mm}
\subsection{Age Progression Synthesis}
\label{APS}
After learning the aging dictionary $\textbf{D}=[\textbf{D}^1,\cdots,\textbf{D}^G]$, for a given face ${\bf x}$ belonging to the age group $g$\footnote{Here, its age range and gender can also be estimated by an age estimator and a gender recognition system in paper \cite{li2015shape}, respectively.}, we can convert it into its aging face sequence $\{{\bf x}^{g+1},...,{\bf x}^{G}\}$. In the aging dictionary learning phase (the offline phase), the neighboring dictionaries are linked via the short-term (i.e., covering two age groups) face aging pairs as the training data. Our aging synthesis (the online phase) should be consistent with this training phase. Therefore, we first generate the aging face ${\bf x}^{g+1}$ in the nearest neighboring age group (i.e., age group $g\!+\!1$) by the learned aging dictionary with one common coefficient, as well as the personalized layer. The coefficient and personalized layer are optimized by solving the
following optimization: 
\vspace{-2mm}
\begin{equation} \label{eq11}
\begin{aligned}
\mathop {\min }\limits_{{{\bf{a}}^g},{{\bf{p}}^g}} &\left\| {\left[ {\begin{array}{*{20}{c}}
		{\bf{x}}\\
		{{{{\bf{x}}}^{g + 1}(0)}}
		\end{array}} \right] - \left[ {\begin{array}{*{20}{c}}
		{{{\bf{W}}^g}}\\
		{{{\bf{W}}^{g + 1}}}
		\end{array}} \right]{{\bf{a}}^g} - \left[ {\begin{array}{*{20}{c}}
		{{{\bf{p}}^g}}\\
		{{{\bf{p}}^g}}
		\end{array}} \right]} \right\|_2^2 \\
&+ \lambda {\left\| {{{\bf{a}}^g}} \right\|_1} + \gamma \left\| {{{\bf{p}}^g}} \right\|_2^2,
\end{aligned}
\vspace{-2mm}
\end{equation}
where ${\bf x}^{g+1}(0)$ is an initial estimation. Eqn.~\eqref{eq11} can be solved by alternatively updating ${\bf{a}}^{g}$ and ${\bf{p}}^{g}$ until convergence, the updating ways are the same as reported in Section~\ref{OP}. After that, taking this new aging face ${\bf x}^{g+1}$ in the current age group $g\!+\!1$ as the input of aging synthesis in the next age group (i.e., age group $g\!+\!2$), we repeat this process until all aging faces have been rendered. Figure~\ref{fig1} shows this age synthesis process.

\begin{algorithm}[t]
	\scriptsize{
		\renewcommand{\algorithmicrequire}{\textbf{Input:}}
		\renewcommand\algorithmicensure {\textbf{Output:} }
		
		\caption{Age Progression Synthesis (Online)}
		\small
		\label{alg2} 
		\begin{algorithmic}[1]
			\REQUIRE  
			{Input face ${\bf x}$ in age range of age group $g$, average faces $\{{\bf r}^g\}_{g=1}^G$, $\lambda$, $\gamma$, and $t\leftarrow1$.}
			\ENSURE{Aging faces ${\bf x}^{g+1}, {\bf x}^{g+2},\cdots,{\bf x}^G$.}
			\renewcommand{\algorithmicrequire}{\textbf{Initialization:}} 
			\REQUIRE 
			{${\bf x}^{g+1}(0)={\bf r}^{g+1}$}, for $g=1,2,\cdots,G-1$.
			\WHILE    {$t<4$}
			\STATE {${\bf x}^{g+1}(t)={\bf x}^{g+1}(t-1)$}, for $g=1,2,\cdots,G-1$.
			\FOR{$c = g,g+1,\cdots, G-1$}
			\STATE Optimize ${\bf \hat{a}}^g$ and ${\bf \hat{p}}^g$ of Eqn.~\eqref{eq11}
			with input pairs $\{{\bf x}^g,{\bf x}^{g+1}(t)\}$.
			\STATE Calculate ${\bf x}^{g+1}(t)={{\bf{H}}^{g + 1}}{{\bf{D}}^{g + 1}}{\bf \hat{a}}^g - {\bf \hat{p}}^g$.
			\ENDFOR
			\STATE $t\leftarrow t+1$.
			\ENDWHILE 
			\STATE {${\bf x}^{g+1}={\bf x}^{g+1}(t-1)$}, for $g=1,2,\cdots,G-1$.
		\end{algorithmic}
	}
\end{algorithm}

\begin{figure}[t]
	\vspace{-2mm}
	\centering
	\subfigure[Convergence curve.]{\includegraphics[scale=0.26]{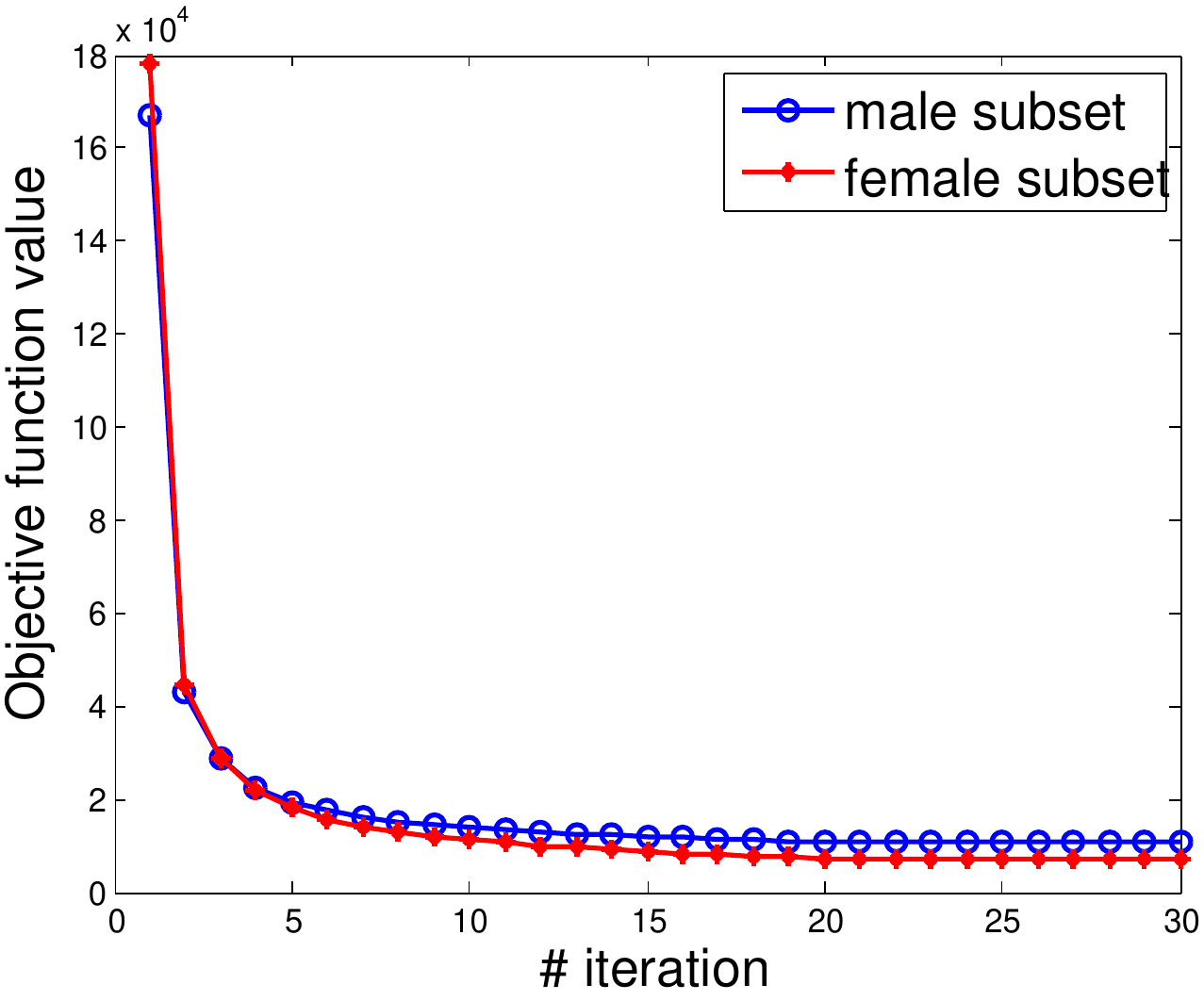}
		\label{fig2a}}
	\subfigure[Three-times aging results.]{\includegraphics[scale=0.21]{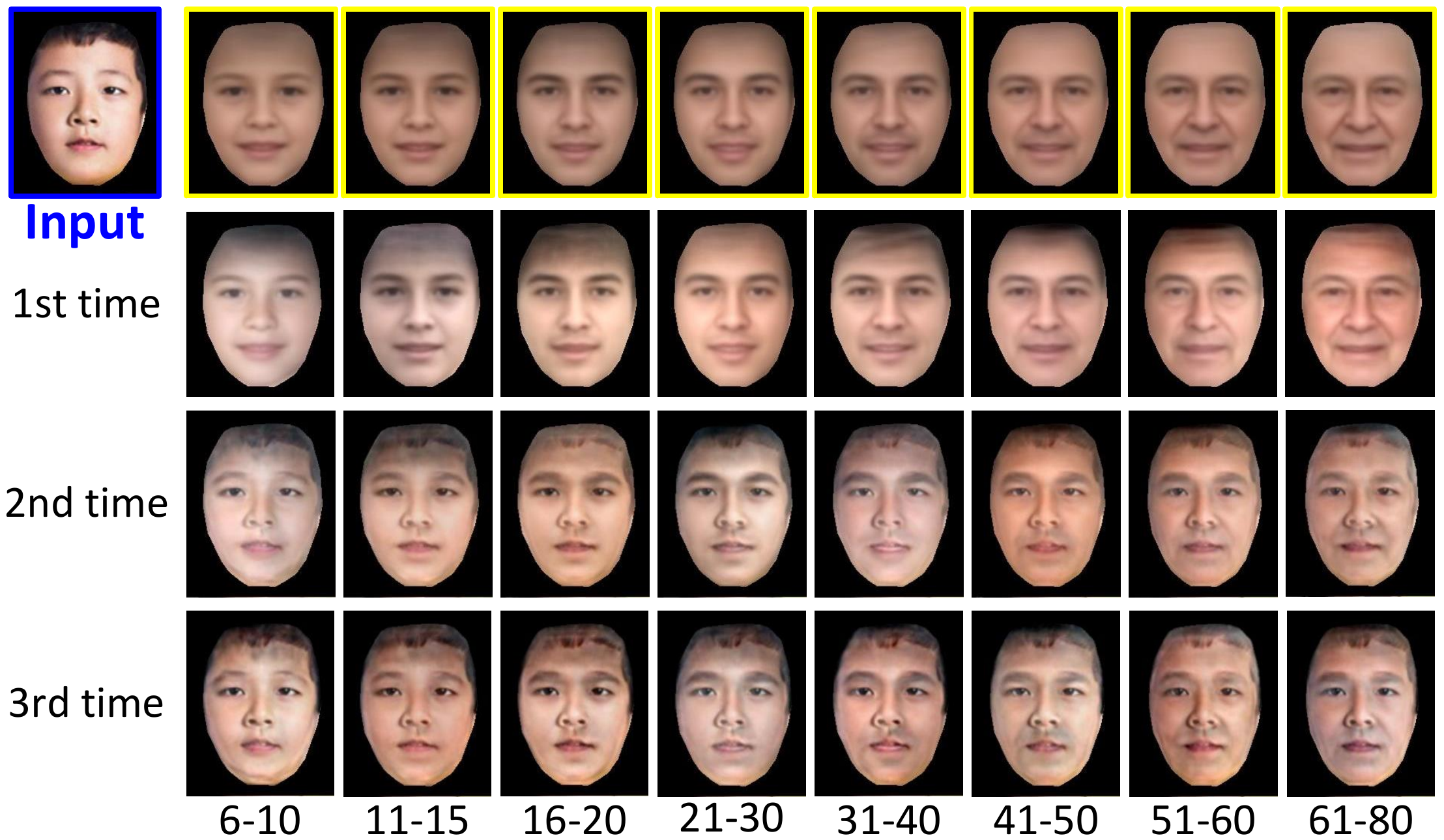}\label{fig2b}}
	\vspace{-3mm}
	\caption{Convergence curve and three-times aging results. (a) Convergence curve of the optimization procedure. (b)~Faces enclosed by the blue-line box and yellow-line box are the input face and average faces, respectively. The aging faces in the 2nd, 3rd and 4th row are the aging outputs of the 1st, 2nd and 3rd time of rendering, respectively. }
	\vspace{-4mm}
\end{figure}

More specifically, if we render an aging face ${\bf x}^{g+1+j}$ for the input ${\bf x}^{g+j}$, an initial setting of ${\bf x}^{g+1+j}$ is needed. In the scenario of age progression, we use the average face ${\bf r}^{g+1+j}$ of the age group $g+1+j$\footnote{In this paper, the average faces are computed by referring to~\cite{kemelmacher2014illumination}.} as the initialization of ${\bf y}^{g+1+i}(0)={\bf r}^{g+1+i}$. However, these outputs ${\bf y}^{g+1}(1)$ for $g=1,2,\cdots,G-1$ are not desired due to the facial differences between individual face and average face. We repeat the rendering of all aging faces with the new input pairs $\{{\bf x}^{g},{\bf x}^{g+1}(1)\}$, $\{{\bf x}^{g+1}(2),{\bf x}^{g+2}(1)\}$,$\cdots$,$\{{\bf x}^{G-1}(2),{\bf x}^{G}(1)\}$. We find that generally we can obtain invariable and desired aging faces when we repeat this process three times. A visualized example is shown in Figure~\ref{fig2b}. Algorithm~\ref{alg2} describes the age progression synthesis in detail.

\section{Experiments}

\subsection{Implementation Details}
\label{ID}
{\bf Data collection.} To train the high-quality aging dictionary, it is crucial to collect sufficient and dense short-time face aging pairs. We download a large number of face photos covering 
different ages of the same persons from Google and Bing image search, and other two available databases, Cross-Age  Celebrity  Dataset  (CACD)~\cite{chen2014cross} and MORPH aging database~\cite{ricanek2006morph}. The CACD database contains more than 160,000 images of 2,000 celebrities with the age ranging from 16 to 62. The MORPH database contains 16,894 face images from 4,664 adults, where the maximum and average age span are 33
and 6.52 years respectively. Both of CACD and MORPH contain quite a number of  short-term intra-person photos. Since these faces are mostly ``in the wild", we select the photos with approximately frontal faces ($- {15^ \circ }$ to ${15^ \circ}$) and relatively natural illumination and expressions. Face alignment~\cite{viola2001rapid} are implemented to obtain aligned faces. To boost the aging relationship between the neighboring aging dictionaries, we use collection flow~\cite{kemelmacher2012collection} to correct all the faces into the common neutral expression. We divide all images into 18 age groups (i.e., $G=9$): 0-5, 6-10, 11-15, 16-20, 21-30, 31-40, 41-50, 51-60, 61-80 of two genders, and find that no person has aging faces covering all aging groups. Actually, the aging faces of most subjects fall into only one or two age groups (i.e. most persons have face photos covering no more than 20 years).  Therefore, we further select those intra-person face photos which densely fall into two neighboring age groups. Finally, there are 1600 intra-person face pairs for training (800 pairs for males, and 800 pairs for females). Every two neighboring age groups for one gender share 100 face aging pairs of the same persons and each age group, except for the ``0-5" age group and the ``61-80" age group, has 200 face photos. We train two aging dictionaries for male and female, respectively.  

{\bf PCA projection.} Take the male subset as an example. We stack $s$ images in the age group $g$ as columns of a data matrix ${\bf M}^g \!\in\! \mathbb{R}^{f\times s}$, where $s\!=\!100$ for $g\in\{1,9\}$, otherwise $s=200$. The SVD of ${\bf M}^g$ is ${{{\bf M}^g}} = {{{\bf U}^g}}{{{\bf S}^g}}({\bf V}^g)^T$. We define the projected matrix ${\bf H}^g={\bf U}^g(:,1\!:\!m)\in\mathbb{R}^{f \times m}$, where ${\bf U}^g(:,1\!:\!m)$ is truncated to the rank~=~$m$ ($m<f$). We use the same strategy for the female subset.

{\bf Parameter setting.} The parameters $\lambda$ and $\gamma$ in Eqn.~\eqref{eq3} are empirically set as $\lambda=0.01$ and $\gamma=0.1$. The number of bases of each aging dictionary is set as $k=70$. In Figure~\ref{fig2a}, we show the convergence properties of aging dictionary learning for male and female subsets. As expected, the objective function value decreases as the iteration number increases. This demonstrates that Algorithm~\ref{alg1} achieves convergence after about $30$ iterations. 


\begin{figure*}[t]
	\centering
	\includegraphics[scale=0.20]{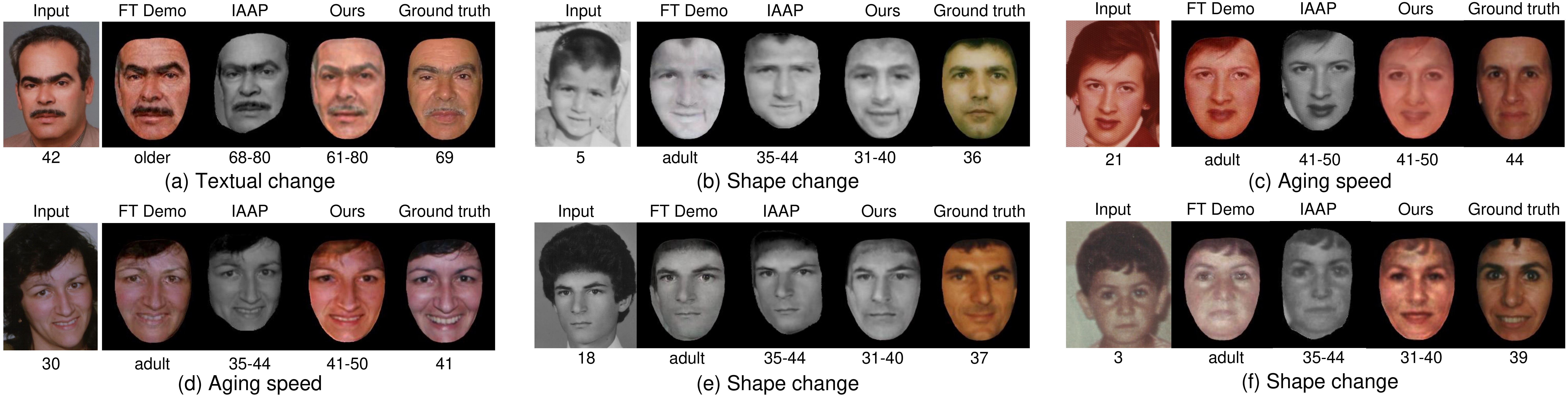}
	\vspace{-1mm}
	\caption{Comparison with ground truth and other methods. Each group includes an input face, a ground truth and three aging results of our method and other two methods. The number or word under each face photo represents the age range (e.g., 61-80) or the age period (e.g., older). For convenience of comparison, black background has been added to each face photo. For better view, please see $\times3$ original color PDF.} 
	\label{fig4}
	\vspace*{-4mm}
\end{figure*}

{\bf Aging evaluation.}
We adopt three strategies to comprehensively evaluate the proposed age progression. First, we qualitatively evaluate the proposed method on the FGNET database~\cite{fgnet}, which is a publicly available database and has been widely used for evaluating face aging methods. This database contains 1,002 images of 82 persons, and the age range spans from 0 to 69: about 64\%
of the images are from children (with ages $<$ 18), and around 36\% are from adults (with ages $\geqslant$ 18). We show the age progression for every photo in the FGNET dataset, and do qualitative comparison with the corresponding ground truth (available older photo) for each person. For reference, we also reproduce some aging results of other representative methods. Second, we conduct user study to test the aging faces of our method compared with the prior works which reported their best aging results. Our method uses the same inputs as in these prior works. Third, cross-age face recognition~\cite{chen2014cross,yadav2013bacteria} and cross-age face verification~\cite{gong2013hidden,wu2012age} are challenging in extreme facial analysis scenarios due to the age gap. A straightforward way for cross-age facial analysis is to use the aging synthesis to normalize the age gap. Specifically, we can render all the faces to their aging faces within the same age range, and then employ the existing algorithms to conduct face verification. Inspired by this, we can also use the face verification algorithm to prove that the pair of aging face and ground truth face (without age gap) is more similar than the original face pair with age gap.

\subsection{Qualitative Comparison with Ground Truth}
We  take  each  photo in FGNET as  the  input  of  our  age  progression. To well illustrate the performance of the proposed age progression, we compare our results with the released results in an  online  fun demo:  Face  Transformer demo (FT Demo)\footnote{http://cherry.dcs.aber.ac.uk/Transformer/}, and also with those by a state-of-the-art age progression method: Illumination-Aware Age
Progression (IAAP)~\cite{kemelmacher2014illumination}. By leveraging thousands
of web photos across age groups, the authors of IAAP
presented a prototyping-based age progression method for automatic age progression of a single photo. FT Demo requires manual location of facial features, while IAAP uses the common aging characteristics of average faces for the age progression of all input faces.  

Some aging examples are given in Figure~\ref{fig4}, covering from baby/childhood/teenager (input) to adult/agedness (output), as well as from adult (input) to agedness (output). By comparing with ground truth, we can see that the aging results of our method look more like the ground truth faces than the aging results of other two methods. In particular, our method can generate personalized aging faces for different individual inputs. In terms of texture change, the aging face of ours in Figure~\ref{fig4}(a) has  white mustache that is closer to ground truth; in shape change, the aging faces of ours in Figure~\ref{fig4}(b)(e)(f) have more approximate facial outline to the ground truth; in aging speed, the faces of FT Demo and IAAP in Figure~\ref{fig4}(c) are aging more slowly, while one of FT Demo in Figure~\ref{fig4}(d) is faster. Overall, the age speed of IAAP is slower than ground truth since IAAP is based on smoothed average faces, which maybe loses some facial textual details, such as freckle, nevus, aging spots, etc. FT Demo performs the worst, especially in shape change. Our aging results in Figure~\ref{fig4} are more similar to the ground truth, which means our method can produce much more personalized results.

\subsection{Quantitative Comparison with Prior Works}

Some prior works on age progression have posted their best face aging results with inputs of different ages, including
\cite{suo2010compositional}, \cite{scherbaum2007prediction}, \cite{ramanathan2009age}, \cite{park2008face}, \cite{patterson2007comparison}, \cite{liang2007age}, \cite{liang2011multi}, \cite{shen2011exemplar}, \cite{sethuram2010hierarchical}, \cite{wang2006age} and \cite{kemelmacher2014illumination}. There are 246 aging results with 72 inputs in total. Our age progression for each input is implemented to generate the aging results with the same ages (ranges) of the posted results. 

We conduct user study to compare our aging results with the published aging results.   
To avoid bias as much as possible,  we invite 50 adult participants  covering a  wide  age range  and  from  all  walks of life. They are asked to observe each comparison group including an input face, and two aging results (named ``A" and ``B") in a random order, and tell which aging face is better in terms of {\em Personality} and {\em Reliability}. {\em Reliability} means the aging face should be natural and authentic in the synthetic age, while {\em Personality} means the aging faces for different inputs should be identity-preserved and diverse. All users are asked to give the comparison of two aging faces using four schemes:  ``A is better", ``B is better", ``comparable", and "neither is accepted", respectively. We convert the results into ratings to quantify the results. 

\begin{figure*}[t]
	\vspace*{-3mm}
	\centering
	\includegraphics[scale=0.22]{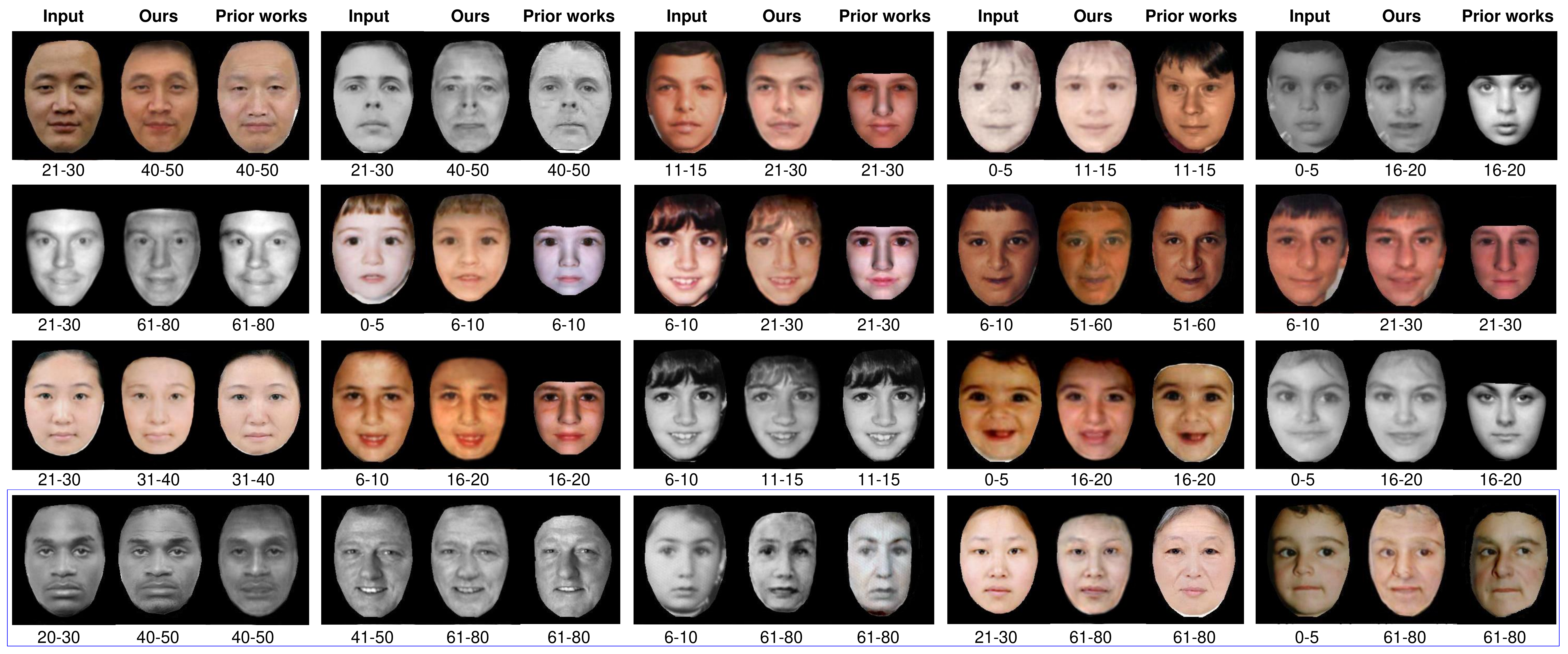}
		\vspace{-6mm}
	\caption{Comparison with prior works. Each group includes an input face and two aging results of ours and prior works. The number under each face photo represents the age range. Some worse results from our method are enclosed by blue box. For convenience of comparison, black background has been added to each face photo. Best viewed in original PDF file.}
	\label{fig5}
	\vspace{-3mm}
\end{figure*}

There are 50 ratings for each comparison, 246 comparison groups, and then 12,300 ratings in total. The voting results are as follows: 45.35\% for ours better; 36.45\% for prior works better; and 18.20\% for ``comparable"; $0$ for ``neither is accepted". The
voting results demonstrate that our method is superior to prior works. 
We also show some comparison groups for voting in Figure \ref{fig5}. Overall, for the input face of a person in any age range, our method and these prior works can generate an authentic and reliable aging face of any older-age range. This is consistent with the gained relatively-high voting support. In particular, for different inputs, our rendered aging faces have more personalized aging characteristics, which further improves the appealing visual sense. For example in Figure \ref{fig5}, the aging faces of ours in the same age range in the $1$st and the $2$nd group of the $1$st row have different aging speeds: the former is obviously slower than the latter; the aging faces of prior works with different inputs in the $1$st and $2$nd groups of the $3$rd column are similar, while our aging results are more diverse for different individual inputs.

\subsection{Evaluation on Cross-Age Face Verification}
\begin{figure}[!t]
	
	\centering
	\subfigure[Pair setting.]{
		\includegraphics[scale=0.222]{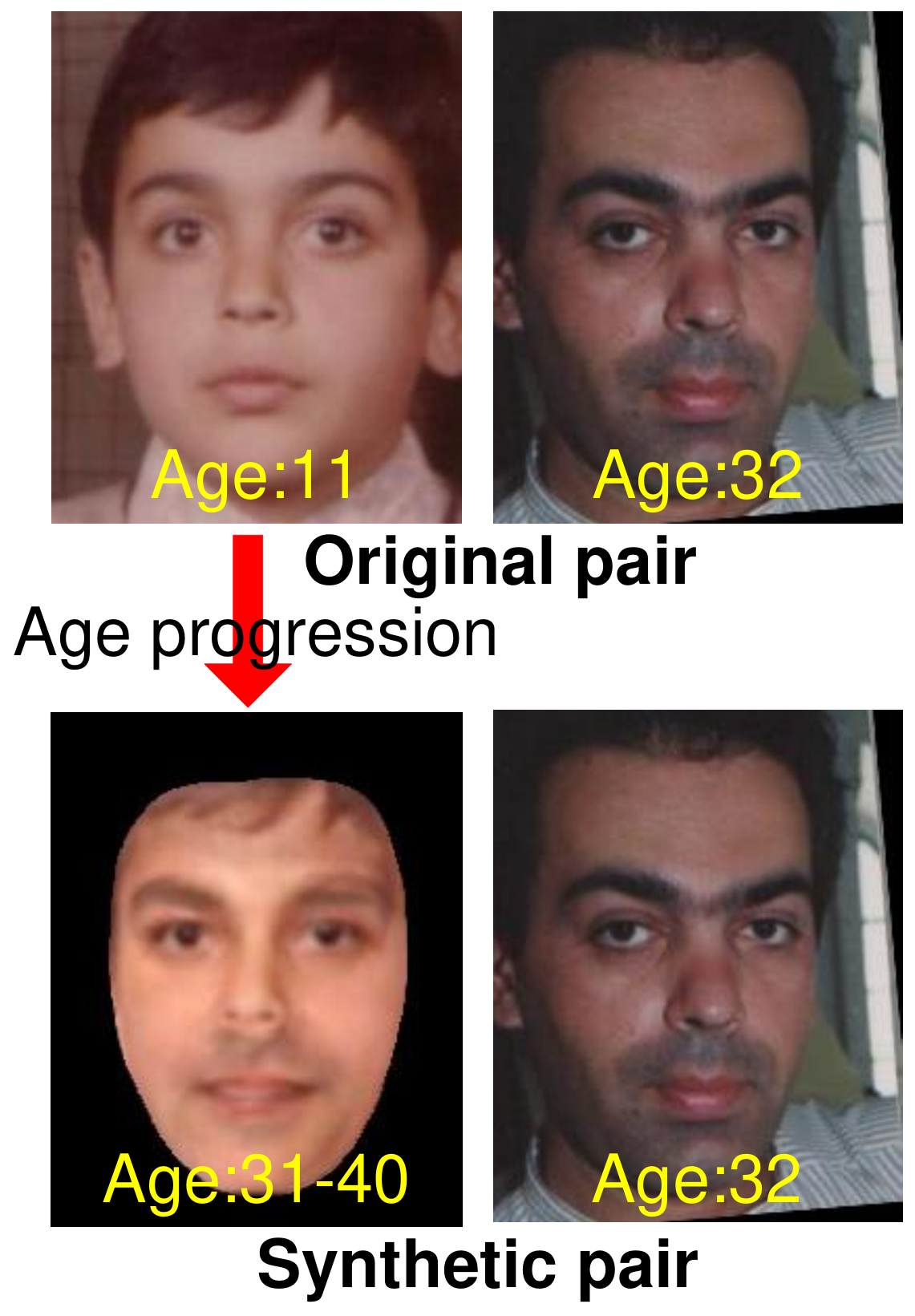}
		\label{fig6a}
	}
	~~~	\subfigure[FAR-FRR curve.]{
		\includegraphics[scale=0.351]{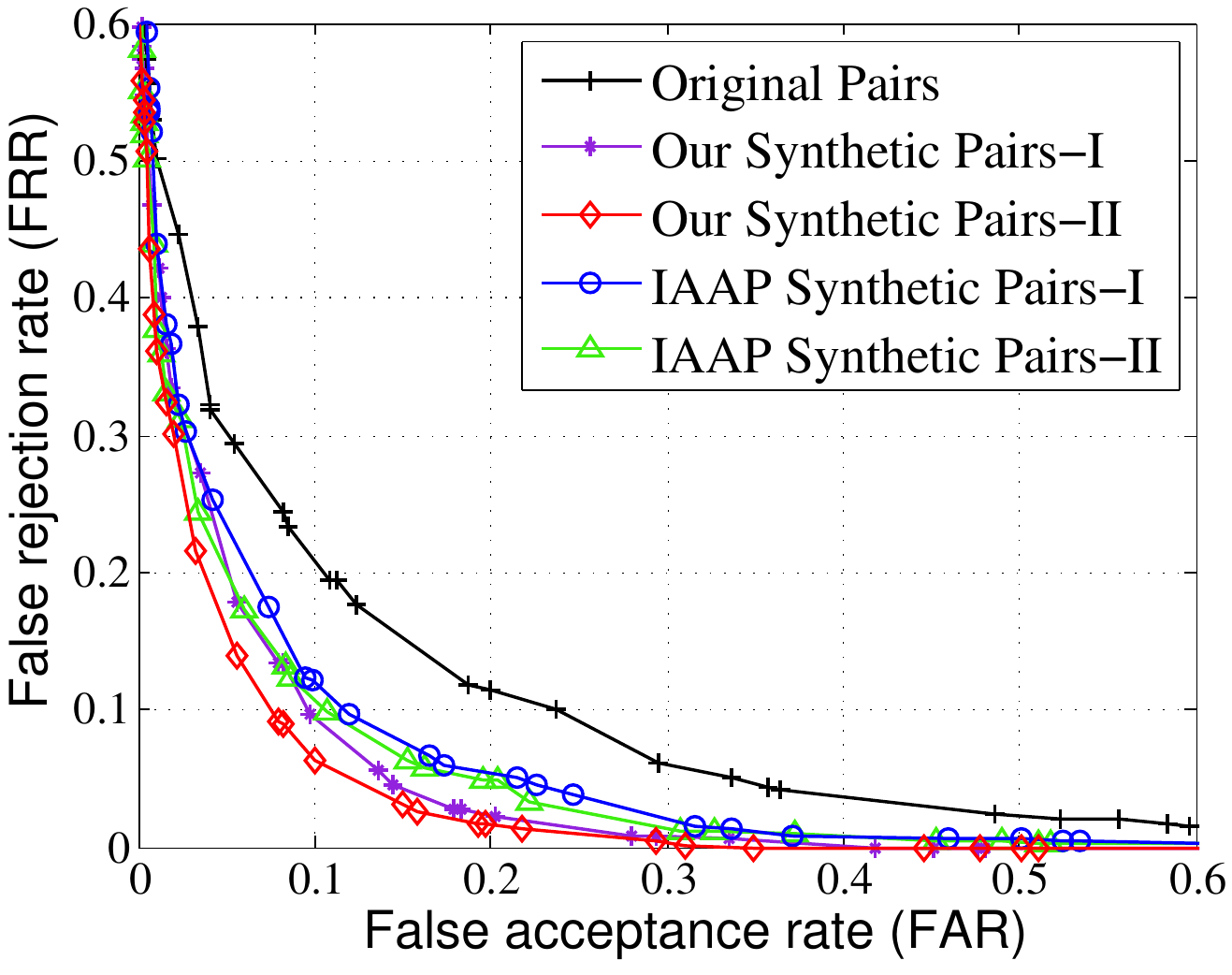} 
		\label{fig6b}
	}
	\vspace{-2mm}
	
	\subfigure[Equal error rates (EER) (\%).]{
		\vspace{-3mm}
		\scriptsize{
			\begin{tabular}{|c|c|c|c|c|c|}	
				\hline			
				\multirow{2}{*}{\hspace{-0.4em}Pair settings\hspace{-0.4em}} &\multirow{2}{*}{\hspace{-0.4em} Original Pairs\hspace{-0.4em}}&\multicolumn{2}{c|}{IAAP Synthetic Pairs}&\multicolumn{2}{c|}{\bf Our Synthetic Pairs} \\
				\cline{3-6}&
				&{\Rmnum{1}}&{\Rmnum{2}}&{\Rmnum{1}}&{\bf \Rmnum{2}}\\
				\hline
				{EER (\%)} &{14.89}&{10.91}&{10.36}&{9.72}&{\bf 8.53}\\
				\hline 	
			\end{tabular}
		}
		\label{fig6c}
	}
	\vspace{-4mm}
	\caption{Pair setting and performance of face verification. Our Synthetic Pairs use our aging synthesis method, while IAAP Synthetic Pairs utilize the IAAP method~\cite{kemelmacher2014illumination}. ``\Rmnum{1}" and ``\Rmnum{2}" denote using actual age and estimated age, respectively.}
	\label{fig6}
	\vspace{-1.5mm}
\end{figure}

To validate the improved performance of cross-age face verification with the help of the proposed age progression, we prepare the intra-person pairs and inter-person pairs with cross ages on the FGNET database. By removing undetected face photos and face pairs with age span no more than 20 years, we select 1,832 pairs (916 intra-person pairs and 916 inter-person pairs), called ``Original Pairs". Among the 1,832 pairs, we render the younger face in each pair to the aging face with the same age of the older face by our age progression method. Replacing each younger face with the corresponding aging face, we newly construct 1,832 pairs of aging face and older face, called ``Our Synthetic Pairs". For fair comparison, we further define ``Our Synthetic Pairs-\Rmnum{1}" as using the given tag labels of FGNET, while ``Our Synthetic Pairs-\Rmnum{2}" is using the estimated gender and age from a facial trait recognition system~\cite{li2015shape}.  To evaluate the performance of our age progression, we also prepare the ``IAAP Synthetic Pairs-\Rmnum{1}" and ``IAAP Synthetic Pairs-\Rmnum{2}" by the state-of-the-art age progression method in~\cite{kemelmacher2014illumination}.
Figure~\ref{fig6a} plots the pair setting. 

The detailed implementation of face verification is given as follows. First, we formulate a face verification model with deep Convolutional Neural Networks (deep ConvNets), which is based on the DeepID2 algorithm~\cite{sun2014deepID2}. Since we focus on the age progression in this paper, please refer to~\cite{sun2014deepID2,taigman2014deepface} for more details of face verification with deep ConvNets. Second, we train our face verification model on the LFW database~\cite{LFWTech}, which is designed for face verification. Third, we test the face verification on Original Pairs, IAAP Synthetic Pairs and Our Synthetic Pairs, respectively.

The false acceptance rate-false rejection rate (FAR-FRR) curves and the equal error rates (EER) on original pairs and synthetic pairs are shown in Figure~\ref{fig6}. We can see that the face verification on Our Synthetic Pairs achieves lower ERR than on Original Pairs and IAAP Synthetic Pairs. This illustrates that the aging faces by our method can effectively mitigate the effect of age gap in cross-age face verification. The results also validate that, for an given input face, our method can render a personalized and authentic aging face closer to the ground truth than the IAAP method. Since the estimated age for an individual is more consistent with human aging tendency, Our/IAAP Synthetic Pairs-\Rmnum{2} outperforms Our/IAAP Synthetic Pairs-\Rmnum{1}.   


\vspace{-1mm}
\section{Conclusions and Future Work}
\vspace{-1mm}
In this paper, we proposed a personalized age progression method. Basically, we design multiple aging dictionaries for different age groups, in which the aging bases from different dictionaries form a particular aging process pattern across different age groups, and a linear combination of these patterns expresses a particular aging process. Moreover, we define the aging layer and the personalized layer for an individual to capture the aging characteristics and the personalized characteristics, respectively. We simultaneously train all aging dictionaries on the collected short-term aging database. Specifically, in two arbitrary neighboring age groups, the younger- and older-age face pairs of the same persons are used to train coupled aging dictionaries with the common sparse coefficients, excluding the specific personalized layer. For an input face, we render the personalized aging face sequence from the current age to the future age step by step on the learned aging dictionaries. In future work, we consider utilizing the bilevel optimization for the personality-aware coupled dictionary learning model.
\vspace{-1mm}
\section{Acknowledgments}
\vspace{-1mm}
This work was partially supported by the 973 Program of China (Project
No. 2014CB347600), the National Natural Science Foundation of
China (Grant No. 61522203 and 61402228), the Program for New Century Excellent
Talents in University under Grant NCET-12-0632 and the
Natural Science Fund for Distinguished Young Scholars of Jiangsu
Province under Grant BK2012033.
\small{
	\bibliographystyle{ieee}
	\bibliography{iccv_camera_ready_shu}
}

\end{document}